\documentclass{article}
\usepackage{spconf,amsmath,graphicx}

\usepackage{enumitem}
\setlist{nosep, leftmargin=14pt}
\usepackage{xcolor}
\usepackage{amsmath,amssymb,amsfonts}

\usepackage{mwe} 
\usepackage{url}
\usepackage{hyperref}

\title{
A Geometric Multimodal Foundation Model Integrating Bp-MRI and Clinical Reports in Prostate Cancer Classification
}

\name{
Juan A. Olmos $^{\star \dagger}$ \qquad Antoine Manzanera$^{\dagger}$ \qquad Fabio Martínez$^{\star}$}

\address{$^{\star}$ Biomedical Imaging, Vision and Learning Laboratory (BIVL$^2$ab), UIS, Colombia.  \\
    $^{\dagger}$ U2IS, ENSTA, Institut Polytechnique de Paris, France.
}

\begin{document}

%
\maketitle
\begin{abstract}



Prostate cancer (PCa) is one of the most common cancers in men worldwide. Bi-parametric MRI (bp-MRI) and clinical variables are crucial for PCa identification and improving treatment decisions. However, this process is subjective to expert interpretations. Furthermore, most existing computer-aided diagnosis methods focus on imaging-based models, overlooking the clinical context and suffering from data scarcity, limiting their ability to learn robust representations. 
We propose a geometric multimodal Foundation Model (FM), named \textit{MFM-Geom}, that learns representations from bp-MRI and clinical reports, encoding visual findings and information from the context of clinical variables. In the representations classification head, the approach leverages symmetric positive definite (SPD) matrices and Riemannian deep learning to integrate imaging-text representations from a biomedical multimodal FM.
Using 10\% of the training data, \textit{MFM-Geom} outperformed baseline class token embedding-based classification (+8.3\%, AUC-PR of 90.67). Generalization on external dataset confirmed the robustness of fine-tuning biomedical FM, achieving an AUC-PR of 90.6.

\end{abstract}
\begin{keywords}
Prostate Cancer, Multimodal, Foundation model, Classification head, Geometric learning
\end{keywords}

\section{Introduction} \label{sec:intro}

Prostate cancer (PCa) is the most commonly diagnosed cancer among men in more than half of countries worldwide \cite{bray2024globalCanStats}. Timely and accurate classification of clinically significant prostate cancer (csPCa) is essential to 
improve treatment outcomes \cite{thestrup-bpVSmp}. Biparametric MRI (bp-MRI), 
has emerged as a valuable imaging modality for the visual assessment of PCa risk, enhancing csPCa detection and reducing unnecessary medical interventions \cite{thestrup-bpVSmp}. Nevertheless, expert evaluation 
remains time-consuming and subject to inter-reader variability.

Deep learning (DL) models have emerged to support PCa detection, predominantly relying on unimodal approximations based on bp-MRI, neglecting complementary diagnostic information \cite{eau_guidelines_2025,saha-picai}. 
For instance, clinical variables are used in PCa screening as preliminary risk indicators in clinical workflows. They provide physicians with a clinical context, suggesting their inclusion in 
DL models to enhance their diagnostic performance.
A recent work proposed a combination of DL–based imaging predictions and clinical variables using machine learning classifiers to differentiate non-csPCa cases from csPCa cases \cite{roest2023multimodalBosma}. However, this decision-level fusion approach overlooks the potential early and complex correlations between imaging-derived information and clinical features that could enhance 
classification. 
This modeling also restricts the development of a deep multimodal representation space. For this, the learning model should provide an embedded (latent) space that is common to imaging and clinical variables while considering the constraints of limited data availability. Multimodal Foundation Models (FM) have shown promising performance, leveraging pretraining on large-scale paired medical datasets \cite{zhang2023biomedclip}. However, adapting these models to specific clinical tasks requires the effective integration of imaging features and key information from clinical features, which can vary significantly across contexts.

This work proposes a novel geometric multimodal foundation model (\textit{MFM-Geom}) that exploits information from bp-MRI and clinical reports within a 
symmetric positive definite (SPD) descriptor and a deep Riemannian learning module to support the classification of csPCa cases. Our contributions are:
(i)~Adaptation of multimodal biomedical FM for PCa diagnosis. The imaging encoder independently analyzes bp-MRI Volumes of Interest (VoIs), and the text encoder includes reports built from clinical variables, including prostate zone location and extra prostatic extension. 
(ii)~A SPD descriptor that captures multimodal correlations from encoders' embeddings, offering an alternative to class token embedding-based classification.
(iii)~Geometric head, a deep Riemannian learning module for processing the SPD descriptor while preserving the geometric structure of the SPD 
manifold. 
(iv)~A contrastive learning loss to align PCa-related imaging and text embeddings into a multimodal latent space. 
Code available at 
\href{https://gitlab.com/bivl2ab/research/publications/2026-ISBI-MFMGeom}{https://gitlab.com/bivl2ab/research/publications/2026-ISBI-MFMGeom}.

\vspace{-2mm}
\section{Proposed Method}

\begin{figure*}[!h]
    \centering
    \includegraphics[width=0.89 \linewidth]{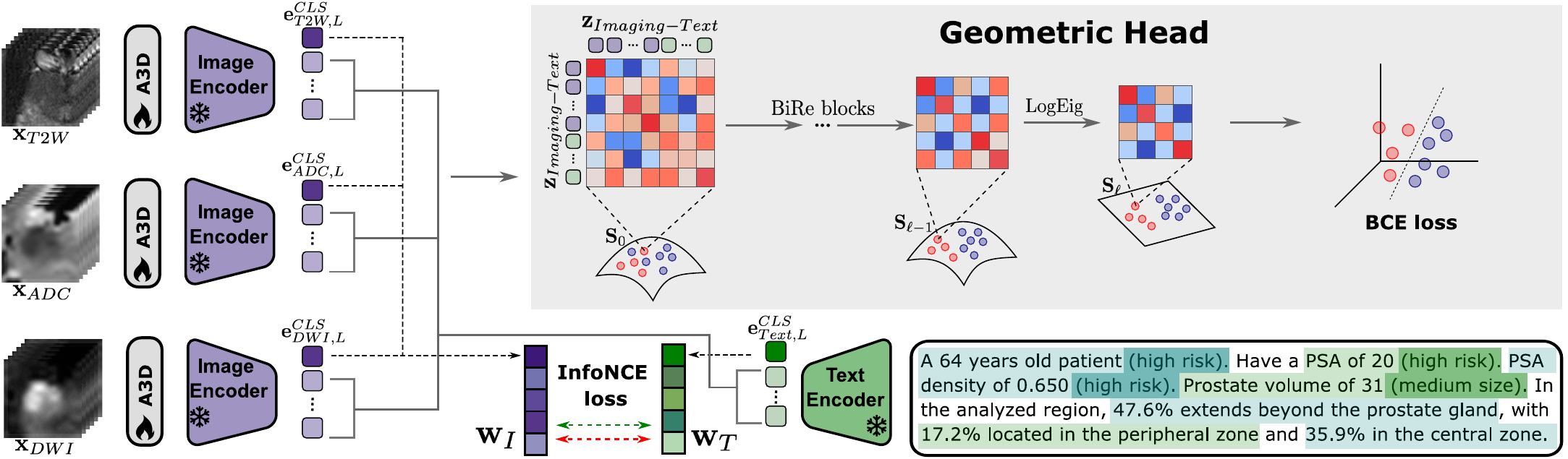}
    \caption{
    \textbf{Pipeline of the proposed geometric multimodal foundation model (\textit{MFM-Geom}).}
    Bp-MRI volumes feed independent image encoders, while clinical variables are converted into a clinical report and fed to the text encoder. The model is fine-tuned using contrastive (InfoNCE) loss on image-text class embeddings, while patch and text embeddings are used to construct a geometric SPD descriptor for classification, considering the Riemannian geometry of SPD matrices space.
    }
    \label{fig:pipeline}
\end{figure*}

The pipeline of the proposed method is presented in Figure~\ref{fig:pipeline}.

\textbf{Bp-MRI representation.}  
This work implements independent volumetric branches to process each bp-MRI sequence
using
the imaging encoder's pretrained vision transformer (ViT) from BiomedCLIP FM was adapted for volumetric inputs using a weight-inflation strategy \cite{solovyev20223d_weights}.  
This volumetric modeling enables the application of $d$ kernels of size $P\times P \times D$ with a stride of $P$. Thus, input VoI $\textbf{x}_* \in \mathbb{R}^{H\times W\times D}$ ($*\in\{T2W,ADC,DWI\}$)
is processed into 3D patches of shape $(P,P,D)$, resulting in 
a sequence with $N_I$ patch embeddings, each of dimension $d$, $\left\{\textbf{e}_{*,0}^i\right\}_{i=1}^{N_I}$.
We denote this 3D patch projection layer as \textit{Adapter 3D} (A3D). In ViTs, a learnable class embedding $\textbf{e}_{*,0}^{CLS}$ is concatenated to this sequence, and a positional embedding is added to form the initial sequence $\textbf{z}_{*,0} = \left[ \textbf{e}_{*,0}^{CLS}, \textbf{e}_{*,0}^{1}, ..., \textbf{e}_{*,0}^{N_I} \right]$. After $L$ attention blocks, ViT outputs a refined sequence $\textbf{z}_{*,L}$. Typically, a classification head, comprising an MLP, is attached to the class embedding $\textbf{z}_{*,L}^{CLS}$. However, this approximation ignores the information learned in the patch embeddings $\{ \textbf{z}_{*,L}^{j} \}_{j=1}^{N_I}$.

\textbf{Clinical report representation.}
A text report $\textbf{x}_{Text}$ was built from clinical variables of the PI-CAI dataset \cite{saha-picai} in a fill-in-the-blank manner, as shown in Fig.~\ref{fig:pipeline}. This textual format conversion adapts to the BiomedCLIP FM's text encoder input and leverages its contextual biomedical knowledge. 
The following variables were included in the report: Age, PSA, PSA density (PSAD), and prostate volume (PV). Complementary variables were derived using prostate zones and gland delineations. A 15 mm-radius sphere centered at each VoI centroid was used to compute its intersection with the peripheral ($P_{pz}$) and central ($P_{cz}$) zones, and its extension outside the gland ($P_{out}$). Based on clinical guidelines \cite{eau_guidelines_2025}, cutoff values were established to categorize the variables: 50 and 60 for age; 10 and 20 for PSA; 0.10, 0.15, and 0.20 for PSAD; 30 and 60 for PV. 
The text report is tokenized into $N_T$ tokens, projected into $d$-dimensional embeddings $\left\{\textbf{e}_{Text,0}^{i} \right\}_{i=1}^{N_T}$, and input to the text encoder. After $L$ attention blocks, it outputs the sequence $\textbf{z}_{Text,L} = \left[ \textbf{e}_{Text,L}^{1}, ..., \textbf{e}_{Text,L}^{N_T}\right]$, where the first embedding corresponds to the class embedding, and the remaining ones correspond to the text tokens embeddings. 

\begin{figure*}[ht]
    \centering
    \includegraphics[width=0.89\linewidth]{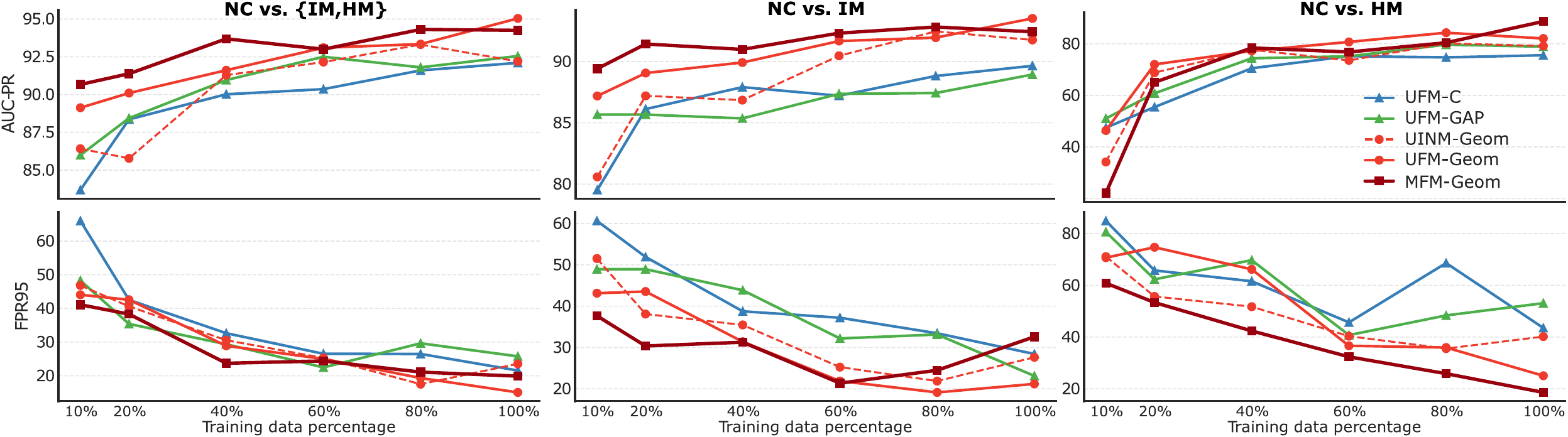}
    \caption{
    \textbf{Classification results} varying training percentage for the three tasks,
    compared to different baselines.
    }
    \label{fig:classification_results}
\end{figure*}

\textbf{Geometric classification head.}
To classify csPCa, we unified the bp-MRI imaging patch and text token embeddings into a multimodal sequence of 
$N$ $= 3N_I + N_T -1$
embeddings, denoted as $\textbf{z}_{Imaging-Text}$. 
Then, a geometric SPD descriptor was calculated to summarize pairwise relationships between the embeddings. For this, we re-arranged $\textbf{z}_{Imaging-Text}$ as a matrix $\textbf{M}\in\mathbb{R}^{N \times d}$ and then compute the SPD descriptor as $\textbf{S}_0 = \frac{1}{d^2}\textbf{M}\textbf{M}^{\intercal}$.  Here, $\textbf{S}_0 (i,j)$ is a similarity measure 
between
the $i$-th and $j$-th embeddings, representing imaging-variables correlations. Since $\textbf{S}_0$ is SPD, it lies on a Riemannian manifold and requires geometry-aware modeling to preserve its intrinsic structure \cite{huang2017_spdnet}. Subsequent geometric learning was performed using \textit{SPDnet} \cite{huang2017_spdnet}. It is composed of a bilinear mapping (\textit{BiMap}) layer that computes a more compact SPD matrix as $\textbf{S}_{\ell} = \textbf{W}_{\ell} S_{\ell-1} \textbf{W}_{\ell}^{\intercal}$, where $\textbf{W}_{\ell} \in \mathbb{R}*^{d_{\ell}\times d_{\ell-1}}$. 
A rectification (\textit{ReEig}) layer that maintains positivity using a threshold $\varepsilon$ via $\textbf{S}_{\ell} = \textbf{U}_{\ell-1} \max(\varepsilon \textbf{I}, \boldsymbol{\Sigma}_{\ell-1}) \textbf{U}_{\ell-1}^{\intercal}$, where $\textbf{S}_{\ell-1} = \textbf{U}_{\ell-1} \boldsymbol{\Sigma}_{\ell-1} \textbf{U}_{\ell-1}^{\intercal}$ is its eigen-decomposition. 
Together, these layers (termed \textit{BiRe} block) project lower-dimensional SPD representations, enhancing multimodal relationships learning to improve csPCa classification.
Finally, SPD representations are projected onto the Euclidean domain via \textit{LogEig} layer, $\textbf{S}_{\ell} = \textbf{log} (\textbf{S}_{\ell-1}) = \textbf{U}_{\ell-1} \log(\boldsymbol{\Sigma}_{\ell-1}) \textbf{U}_{\ell-1}^{\intercal}$. The resulting representations were then passed through an MLP for classification.

The multimodal BiomedCLIP FM is pretrained using contrastive learning between millions of (image, text) pairs \cite{zhang2023biomedclip}. To leverage this multimodal latent space, we propose fine-tuning our model
using an image embedding $\textbf{w}_I = \mathrm{MLP}\left( [\textbf{z}_{T2W,L}^{CLS}, \textbf{z}_{ADC,L}^{CLS}, \textbf{z}_{DWI,L}^{CLS} ] \right)$ from BP-MRI class embeddings, and text embedding $\textbf{w}_T = \mathrm{MLP} \left( \textbf{z}_{Text,L}^{CLS} \right)$. These projections are aligned in a shared latent space, where the similarity between corresponding imaging-text pairs is maximized and non-corresponding pairs minimized.

\vspace{-3mm}
\subsection{Datasets}
We used the PI-CAI dataset \cite{saha-picai}. Each bp-MRI volume was standardized 
to a size of $384 \times 384 \times 24$ with a voxel spacing of $0.5 \times 0.5 \times 3.0~\text{mm}^3$, and normalized to $[0,1]$. Each study was stratified by histopathological verification using ISUP grade \cite{saha-picai}. We used 415 csPCa lesions (ISUP $\ge 2$): 240 ISUP-2, 91 ISUP-3, 37 ISUP-4, and 47 ISUP-5,  
for which
VoIs of size (64,64,8) were extracted around the centroid of expert delineations. 
Additionally, the dataset comprises 847 studies without csPCa lesion, for which
a single VoI of the same size
was extracted in a 
sextant randomly drawn 
according to 
the sextant-wise spatial
distribution of csPCa lesions.
To evaluate generalization, we considered the external dataset \textit{PROSTATE158} \cite{adams2022prostate158}, which includes 102 csPCa and 56 non-csPCa cases.
The VoIs were constructed from these cases using the same methodology.

\vspace{-2mm}
\subsection{Experimental Setup}

We employed BiomedCLIP FM, pretrained on 15 million image–text pairs extracted from biomedical articles in PubMed Central \cite{zhang2023biomedclip}. The image encoder is a ViT-B/16. The text encoder is PubMedBERT, also pretrained on PubMed,  comprising a domain-specific vocabulary.
In the geometric head, we used two \textit{BiRe} blocks.
For fine-tuning, the loss function is the sum of the binary cross-entropy and 
contrastive (InfoNCE) loss.  
We used an AdamW optimizer with a learning rate of $10^{-4}$ for non-SPD layers, and a 
geometry-aware 
algorithm with a learning rate of $10^{-2}$ for SPD layers \cite{huang2017_spdnet}. 

We made classification experiments between non-csPCa (NC), intermediate (IM; ISUP 2–3), and high malignancy (HM; ISUP 4–5) cases in three tasks: NC {\em vs} \{IM, HM\}, NC {\em vs} IM, and NC {\em vs} HM. We employed a 5-fold 
cross-validation scheme \cite{saha-picai}.
Due to data imbalance, we evaluated AUC-PR, and FPR95 (false positive rate at 95\% true positive rate) score. Statistical significance was assessed using a two-sided Wilcoxon signed-rank test applied to paired model predictions and for each metric. 
We first evaluated the unimodal (imaging-only) FM model (\textit{UFM}) configuration to validate the proposed geometric modeling
within the ViT BiomedCLIP image encoder (\textit{UFM-Geom}). 
Two baseline methods were considered: 
(i)~a standard classification head using bp-MRI class embeddings and an MLP for classification (\textit{UFM-C}), and 
(ii)~a global average pooling approach over all bp-MRI patch embeddings followed by an MLP for classification (\textit{UFM-GAP}). 
To assess the contribution of BiomedCLIP pretraining, we considered the same ViT but pretrained on ImageNet (\textit{UINM-Geom}). Finally, we included the proposed multimodal geometric model (\textit{MFM-Geom}).

\section{Evaluation and Results}

To evaluate learning representation robustness, we assessed models’ performance under reduced training data (see Figure~\ref{fig:classification_results}). Models with the geometric module achieved the best results across all classification scenarios, in first position \textit{MFM-Geom}, followed by \textit{UFM-Geom} and \textit{UINM-Geom}.
In the NC vs. \{IM, HM\} task, using 10\% of the training set, \textit{MFM-Geom}
achieved $90.67 \pm 1.17$, outperforming all unimodal configurations by 7.0 (+8.3\%), 4.7 (+5.4\%), 4.3 (+4.9\%), and 1.5 (+1.7\%) points over \textit{UFM-C}, \textit{UFM-GAP}, \textit{UINM-Geom}, and \textit{UFM-Geom}, respectively ($p < 0.05$). 
In FPR95 it scores $41.10 \pm 5.06$, 24.9 (–37.7\%), 7.2 (–14.9\%), 5.7 (–12.2\%), and 2.9 (–6.7\%) points less, relative to the same unimodal models ($p < 0.05$).
Using the entire dataset, state-of-the-art CNN-based approaches report AUC-ROC scores between $94.1$ and $96.5$, whereas the proposed \textit{UFM-Geom} and \textit{MFM-Geom}
achieved $97.2$ and $96.7$ respectively, demonstrating competitive performance \cite{2025geometricnca}.
In the NC vs. IM task, 
the proposed geometric models \textit{MFM-Geom}, and \textit{UFM-Geom} provided a greater advantage over baseline models and 
\textit{UINM-Geom}
underscoring the benefits of geometric representation learning and supporting the use of biomedical FM encoders. With only 10\% of the training data, \textit{MFM-Geom} achieved an AUC-PR of $89.4 \pm 1.5$, exceeding \textit{UFM-Geom} and \textit{UFM-C} by 2.2 (+2.6\%) and 9.9 (+12\%), respectively ($p < 0.05$).
In the NC vs. HM task, performance differences across training data percentages were marginal,
likely due to the class imbalance present in this task.
Nevertheless, a consistent advantage in FPR95 is observed for \textit{MFM-geom} across all training-data percentages.

\begin{figure}[!h]
    \centering
    \includegraphics[width=0.75\linewidth]{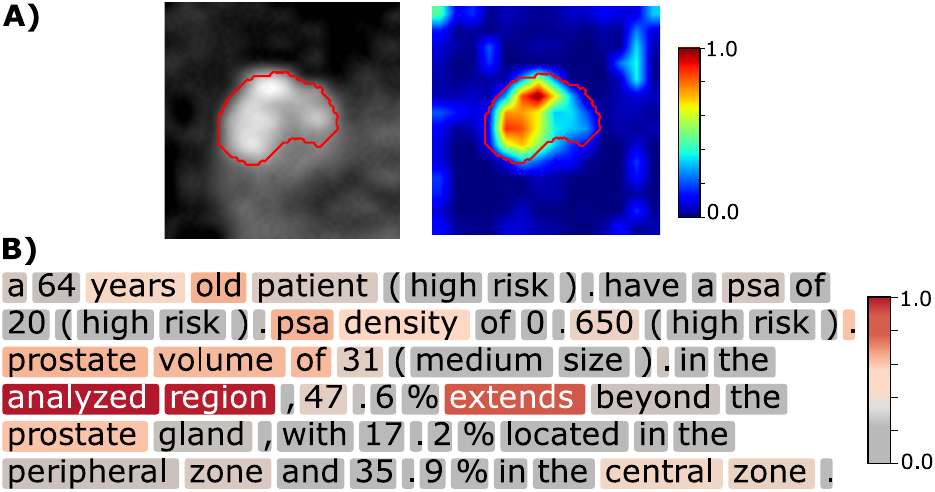}
    \caption{\textbf{Attention maps} of the proposed method 
    extracted
    from the image (A) and text (B) encoders.}
    \label{fig:explainability}
\end{figure}

One limitation of FM is their limited interpretability. 
We show some attention maps of the proposed \textit{MFM-Geom} in Figure~\ref{fig:explainability}. They are computed from the mean attention matrix of the Image/Text encoder's last block, using attention coefficients between the class token and image/text tokens.
The image encoder focused on lesion, while the text encoder highlighted variables such as PSAD, PV, lesion extension beyond the gland, and prostate zone. This pattern, observed across multiple cases, 
\textcolor{black}{
underscores the efficacy of the proposed method in learning multimodal representations 
coherent with PCa-related imaging and textual findings.
}

In the external validation, 
\textit{UFM-Geom}
outperformed the baseline models, highlighting the effectiveness of the geometric head in improving generalization (see Table~\ref{tab:predictions_p158}). 
Furthermore, these results underscore the importance of fine-tuning a FM pretrained on biomedical data, as the \textit{UINM-Geom} model demonstrates generalization performance comparable to baseline methods.
As this dataset lacks clinical variables, multimodal setting was not evaluated.

\begin{table}[!h]
\centering
\resizebox{0.27 \textwidth}{!}{
\begin{tabular}{lcc} 
\hline
\textbf{Method} &  \textbf{AUC-PR $\uparrow$} & \textbf{FPR95 $\downarrow$} \\ \hline
\textit{UFM-C} &  86.52 & 53.15 \\
\textit{UFM-GAP} &  85.26 & 44.59 \\
\textit{UINM-Geom} & 84.80 & 54.50 \\
\textbf{\textit{UFM-Geom}} & \textbf{90.61} & \textbf{37.39} \\ \hline
\end{tabular}}
\caption{
\textbf{
Classification
performance} in the external cohort.
}
\label{tab:predictions_p158}
\end{table}

\textcolor{black}{
The proposed design enabled models with a reasonable number of parameters (in M): 8.4, 6.3, and 6.7 for \textit{MFM-Geom}, \textit{UFM-Geom}, and \textit{UFM-C}. 
The
geometric head increased the inference time per sample (in ms): 14.3, 12.3, and 7.6 for \textit{MFM-Geom}, \textit{UFM-Geom}, and \textit{UFM-C}, respectively.
}

\section{Conclusions}
This work introduced a novel geometric multimodal foundation model (\textit{MFM-Geom}), enabling more efficient joint learning of imaging and clinical report findings for the classification of csPCa. 
The proposed geometric head outperformed traditional classification heads for FM imaging and ImageNet-pretrained encoders. 
The learned geometric multimodal representations improved performance over traditional classification heads for FM imaging under limited data conditions, particularly in differentiating intermediate-malignancy lesions, highlighting their potential clinical relevance in enhancing the detection of cases suitable for active surveillance and improving treatment-decision accuracy. 
Future work includes an evaluation of the multimodal geometric latent space and its capability to stratify different levels of malignancy. 
 
\section{Compliance with ethical standards}
This work used the public PI-CAI dataset \cite{saha-picai}. Ethical approval was not required per the open-access dataset license.



{
\bibliographystyle{IEEEbib}
\bibliography{refs}
}


\end{document}